\pdfoutput=1

\documentclass[11pt]{article}

\usepackage[final]{acl}

\usepackage{times}
\usepackage{latexsym}

\usepackage[T1]{fontenc}

\usepackage[utf8]{inputenc}

\usepackage{microtype}

\usepackage{inconsolata}

\usepackage{graphicx}

\usepackage{booktabs}
\usepackage{caption}
\usepackage{subcaption}
\usepackage{multirow}
\usepackage{comment}
\usepackage[htt]{hyphenat}

\usepackage[most]{tcolorbox}
\tcbset{colback=gray!5,colframe=gray!50!black,arc=2mm,boxrule=0.6pt}
\newtcolorbox{plainbox}[1][]{
  title={#1},
  colback=blue!5,     
  colframe=blue!60,   
  boxrule=0.5pt,      
  arc=2mm,            
  fonttitle=\bfseries,
  breakable           
}

\title{\textsc{TempViz}: On the Evaluation of Temporal Knowledge \\ in Text-to-Image Models}

\author{
  \textbf{Carolin Holtermann\textsuperscript{1}},
  \textbf{Nina Krebs\textsuperscript{2}},
  \textbf{Anne Lauscher\textsuperscript{1}}
\\
  \textsuperscript{1}Trustworthy AI Lab, University of Hamburg \\
  \textsuperscript{2}University of St. Gallen
\\
  \texttt{carolin.holtermann@uni-hamburg.de}
}

\begin{document}
\maketitle
\begin{abstract} 
Time alters the visual appearance of entities in our world, like objects, places, and animals. Thus, for accurately generating contextually-relevant images, knowledge and reasoning about time can be crucial (e.g., for generating a landscape in spring vs. in winter). Yet, although substantial work exists on understanding and improving temporal knowledge in natural language processing, research on how temporal phenomena appear and are handled in text-to-image (T2I) models remains scarce. We address this gap with \textsc{TempViz}, the first data set to holistically evaluate temporal knowledge in image generation, consisting of 7.9k prompts and more than 600 reference images. Using \textsc{TempViz}, we study the capabilities of five T2I models across five temporal knowledge categories. Human evaluation shows that temporal competence is generally weak, with no model exceeding 75\% accuracy across categories. Towards larger-scale studies, we also examine automated evaluation methods, comparing several established approaches against human judgments. However, none of these approaches provides a reliable assessment of temporal cues -- further indicating the pressing need for future research on temporal knowledge in T2I. We publish all data and code at \url{https://github.com/TAI-HAMBURG/TempViz}.

\end{abstract}

\section{Introduction}
\begin{quote}
\textit{``Time is the longest distance between two places.''} 

\vspace{-1.5em}
\begin{flushright}
-- \textit{Tennessee Williams, 1944}
\end{flushright}
\end{quote} 

\noindent An artwork from \emph{1900} looks different than an artwork from \emph{1960}, and a dog at age \emph{3 months} looks different than a dog at age \emph{10 years}. Similarly, landscapes change their appearance across seasons, such as \emph{winter} or \emph{summer} 
-- Time influences the visual appearance of the world around us. 
\begin{figure}
    \centering
    \includegraphics[width=1.0\linewidth]{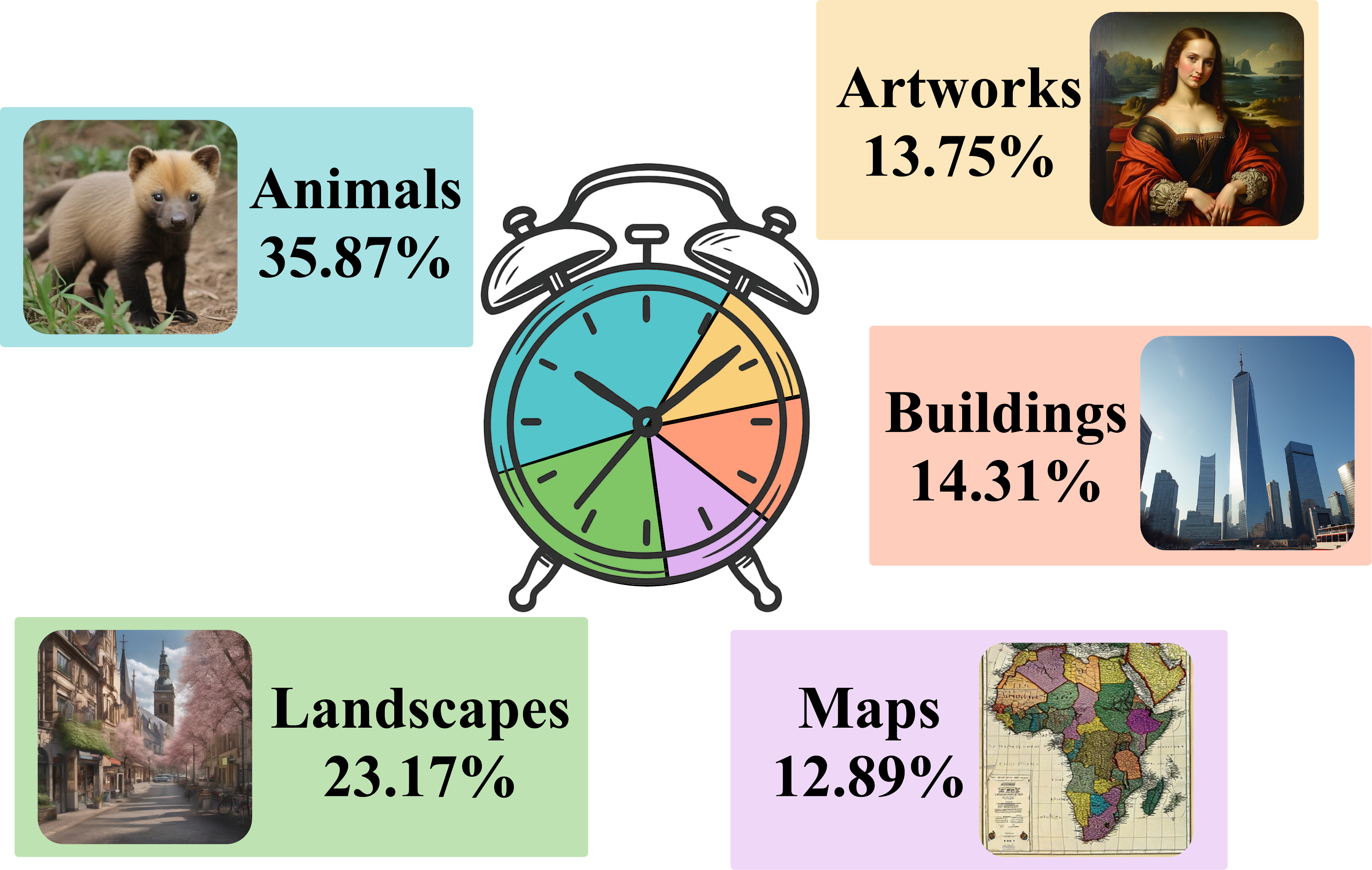}
    \caption{Distribution of the temporal categories within \textsc{TempViz} with example images of different T2I models.}
    \label{fig:tempvizDistribution}
    \vspace{-1em}
\end{figure} 

The effects of time and their perception are intrinsically intertwined with our senses~\citep[e.g.,][]{ma_memorability_2024}, and play a major role in most human societies. For accurately representing the world, an understanding of the temporal effects (e.g., aging) and corresponding visual effects is thus crucial. Users might specify in an input prompt to image generation (e.g., to a Stable Diffusion model~\cite{rombach_high-resolution_2022}) the particular epoch or year they are looking for, expecting the generated output to be aligned with their instruction. 

In natural language processing (NLP), many studies have analyzed temporal knowledge capabilities of NLP systems, for instance, relating to the temporal reasoning capabilities of large language models~\citep[e.g.,][\emph{inter alia}]{gurnee2024languagemodelsrepresentspace, fatemi2024testtimebenchmarkevaluating}.
However, despite the large body of existing works on temporal knowledge in text-only models, and the central role of temporal knowledge and reasoning for accurate visual representations, the landscape on resources and evaluation of temporal knowledge in text-to-image (T2I) models remains largely unexplored.

In this work, we address this research gap by posing the following research question:  \textbf{\emph{How well do existing T2I models perform at following temporal instructions across different subject categories?}} To answer this question, we introduce a novel resource for analyzing temporal knowledge in T2I models and conduct both manual and automatic evaluations. We find that even the strongest T2I model among those evaluated correctly applies the correct temporal visual cues in at most 75\% of cases for only one category, and that for some categories the temporal instruction is barely followed (for example, as low as 15\% for historical maps). Strikingly, none of the automated evaluators we tested was able to reliably judge the correct application of temporal knowledge above a macro F1 of 72\%. Together, these results motivate more targeted research on temporal knowledge in image generation and evaluation.

\paragraph{Contributions} Our contributions are threefold: (\textbf{1})  We present \textsc{TempViz}, the first dataset specifically designed for analyzing temporal knowledge in T2I models. It comprises 7,940 prompts across five diverse categories (e.g., \emph{animals}, \emph{artworks}), each paired with category-appropriate temporal cues (e.g., \emph{lifespans}, \emph{art periods}, \emph{seasons}), and includes reference images that exemplify correct generations for a subset of categories. (\textbf{2}) Using \textsc{TempViz}, we conduct a human evaluation of five recent text-to-image models, measuring image quality, subject fidelity, and adherence to the specified temporal cue. (\textbf{3}) We complement the manual study with an extensive assessment of automatic evaluation methods, analyzing their suitability for judging whether temporal knowledge has been applied correctly in generated images.

\section{Related Work}
We present an overview of datasets for evaluating temporal knowledge in NLP, as well as datasets designed to assess T2I models.

\paragraph{Temporal Knowledge Datasets}
In large language models, temporal knowledge has been examined from multiple perspectives. Early work on temporal relation extraction introduced annotated corpora such as TimeBank \cite{timebank}. Subsequent benchmarks probe time-varying factual knowledge and temporal reasoning, including the TempLAMA benchmark \cite{templama} and its derived datasets such as LAMA-TK \cite{zhao-etal-2022-language} and TempReason \cite{tan-etal-2023-towards}. Other work in the direction of LLM probing focuses rather on time-sensitive question answering \cite{chen2021datasetansweringtimesensitivequestions, holtermann-etal-2025-around}.

Another line of work in the direction of temporal NLI focuses on inferring knowledge about temporal relations and more complex temporal reasoning. This includes the ordering of temporal events \cite{thukral2021probinglanguagemodelsunderstanding, wang2024trambenchmarkingtemporalreasoning} and temporal grounding \cite{liu2023groundingcomplexnaturallanguage, fatemi2024testtimebenchmarkevaluating}. In addition, several studies investigate methods for improving temporal reasoning through specialized training objectives \cite{xiong-etal-2024-large, tan2024robusttemporalreasoninglarge}.

\paragraph{Text-to-Image Evaluation Datasets} Compared to the amount of benchmarks available to evaluate the capabilities of LLMs, the landscape of resources to evaluate T2I models still remains scarce. Most benchmarks focus on the evaluation of the general text-image alignment \citep[e.g.,][]{huang2023t2icompbenchcomprehensivebenchmarkopenworld, yarom2023readimprovingtextimagealignment}. \citet{park2021} specifically focus on the question of how well these models generalize to novel word compositions. \citet{lee2023holisticevaluationtexttoimagemodels} introduce HEIM, a benchmark for the holistic evaluation of T2I models, which includes 12 evaluation aspects such as text-image alignment, image quality, aesthetics, originality, reasoning, knowledge, bias, toxicity, fairness, robustness, multilinguality, and efficiency; still neglecting temporally dependent depictions. Concurrent to our work \citet{niu2025wiseworldknowledgeinformedsemantic} present WISE, a benchmark for assessing world knowledge reasoning in T2I models with 1,000 prompts spanning cultural commonsense, spatio-temporal reasoning, and natural science. While they also cover aspects of temporal knowledge, their focus is rather broad and the scale of their dataset is modest: Of the dataset, only 167 prompts concern temporal knowledge, targeting mostly narrow aspects like period-specific instruments instead of the wider visual manifestations of temporal change.  As a result, WISE provides limited insights for analyzing and improving these fine-grained temporal effects in image generation. 

\section{TempViz}
We analyze how established T2I models interpret and represent temporal cues during image generation. To this end, we introduce \textsc{TempViz}, a carefully curated dataset of 7,940 instructions spanning five subject categories (see Figure \ref{fig:tempvizDistribution}), chosen to reflect common use cases where temporal effects are visually observable. Moreover, the prompts cover multiple dimensions of temporal variation at varying levels of difficulty, enabling a focused assessment of models' temporal reasoning. We intentionally exclude prompts that generate human individuals due to ethical concerns: varying both temporal cues and subjects within this category would necessarily involve prompting different demographic backgrounds, introducing complex issues of stereotype reinforcement and bias that would shift the scope from temporal reasoning to fairness considerations. We therefore chose to omit humans and focus on categories where temporal variation can be analyzed with fewer confounds. To support reliable human annotation, we also collect semantically similar real-world reference images for each prompt and provide concise textual descriptors of the visual attributes expected in the generations, when available. In \S\ref{sec:exp}, we demonstrate how \textsc{TempViz} can be used to evaluate temporal image-text alignment.

\subsection{Creation of Prompts}
We query all models using four paraphrases of the template prompt ``[...] a photorealistic image of \{$s_i$\} in \{$t_i$\}''. Each prompt instantiates a subject $s_i$ from category $c \in S$ and a category-specific temporal cue $t_i$ where $S=$\{\textit{landscapes, animals, artworks, buildings, maps}\}.

\paragraph{Landscapes}
A natural test of temporal knowledge in T2I models is the rendering of seasonal and daily variation in landscapes. Therefore, we select commonly known, generic natural environments that evolve over time (e.g., \textit{a lake}). Next, we define a set of temporal cues $t_n$, including all four seasons (for example \textit{in spring}, \textit{in July}, or \textit{07/17/1899}), and times of the day (e.g., \textit{morning}). To reduce prompt ambiguity in seasonality and daylight, we append the suffix \textit{in Europe} to every prompt and restrict generations to the Northern hemisphere. Correct image generation for this category requires models to integrate multiple knowledge types, specifically commonsense knowledge of seasonal and diurnal cycles, temporal grounding that maps temporal information to seasons and times of day, and geographical knowledge of European climate patterns.

\paragraph{Animals}
Temporal knowledge is also required to depict biological aging. Here, we focus on mammals, as their aging process is more visually recognizable to human observers than that of insects or amphibians. To assemble a representative set of mammals, we prompt GPT-4~\citep{openai_gpt-4_2024} to provide \textit{``a set of well-known mammals that show visible signs of aging''}, then consolidate frequent suggestions, remove duplicates, and exclude extinct species. As temporal cues, we defined four distinct life stages: young (1 month old), adult (50\% of life span), very old (80\% of life span), and well beyond the expected life span (200\% of life span). To this end, we determine the typical lifespans of the selected animals from HAGR\footnote{\url{https://genomics.senescence.info/}}, an age database of animal aging and longevity. This category requires models to combine factual knowledge about how aging manifests visually in different species with temporal grounding that maps age expressions to species-specific lifespans.

\begin{table*}[t!]
\small
    \centering
    \begin{tabular}{llc l}
        \toprule
        \textbf{Category} & \textbf{Selection Methodology} & \textbf{\# Prompts} & \textbf{Prompt Example}\\
        \midrule
        & & & Create a photorealistic image of ...\\
        Landscapes  & 92 dates $\times$ 12 locations & 460 & ... a landscape in january\\ 
        Animals  & 4 lifespans $\times$ 178 mammals  & 712 & ... an Alpaca that is one month old \\
        Buildings  & 4 dates $\times$ 71 buildings & 284 & ... the Berlin Wall in Germany on 10/23/1983\\
        Maps  & 64 dates $\times$ 12 regions & 256 & ... a map of europe in 1938 \\
        Artworks  & 10 dates $\times$ 13 periods $\times$ 2 & \multirow{ 2}{*}{273} & ... an artwork in a style of [Ancient Art | that was popular in  \\
        &         + 13 periods by name    &  & the Western world in 1732]\\
        \midrule
        \textbf{Total} & \textbf{x4 prompt variations} & \textbf{7,940}\\
        \bottomrule
    \end{tabular}
    \caption{Composition of the \textsc{TempViz} dataset across the five subject categories. For \textit{Artworks}, we include two additional category-specific templates.}
    \label{tab:full_dataset_distribution}
    \vspace{-1em}
\end{table*}

\paragraph{Buildings}
Another form of temporal change appears in the built environment. Urban form evolves through events such as war, redevelopment, and natural disasters. As city-scale changes are hard to judge from a single image, individual buildings offer a more tractable unit for human annotation. We therefore focus on buildings and structures that have been destroyed over time (e.g., \textit{the World Trade Center before 2001}). Using Wikidata\footnote{\url{https://www.wikidata.org}}, we identify subjects via SPARQL queries that have known inception and demolition dates, a country label, and a Wikipedia page in at least five languages. As temporal cues, we use four dates: two close to the demolition (just before and just after) and two far from it (well before and well after). To prevent ambiguity from globally recurring building names, we include country information in each prompt. Success in this category requires models to combine event-specific temporal knowledge with causal temporal reasoning (before vs. after the demolition) and factual knowledge about the building's correct appearance at that time.

\paragraph{Maps}
Complementing previous categories, we evaluate temporal knowledge in a cartographic context. Therefore, we first use GPT-4 to identify 12 regions with documented border changes over time. Then, to select salient time points, we query the Google Autosuggest API\footnote{\url{https://github.com/wildlyinaccurate/google-suggestqueries-api}} for frequently searched completions for each region, and then manually validate all resulting timestamps. We list the number of suggestions per region in the Appendix \ref{app:maps_distr}. This category tests factual temporal knowledge of time-stamped geopolitics, geographic reasoning about borders and land shapes, and symbolic rendering conventions for cartographic depiction.

\paragraph{Artworks}
Recent T2I models have demonstrated advanced performance in generating images in well-known artistic styles \cite{podell2023sdxlimprovinglatentdiffusion}, suggesting access to factual knowledge about historical art movements. We thus include an Artworks category to evaluate temporal and period knowledge. Our premise is that, given models' ability to imitate specific artistic styles, remaining errors primarily reflect weak temporal grounding. Again, we first prompt GPT-4 to propose popular styles (e.g., \textit{Impressionism}), then manually assign the period during which each style was prominent (e.g., \textit{1863--1883}). In this setting, the subject and the temporal cue jointly define the style. For each style, we sample multiple dates within its period and, unlike the other categories, employ two templates: one that names the period explicitly and one that specifies a concrete date within it. To reduce ambiguity and simplify annotation, we restrict prompts to Western art styles and choose dates that lie clearly within, rather than near the boundaries of well-known art epochs. This category requires models to ground timestamps in the correct stylistic period and apply stylistic pattern matching, requiring both temporal reasoning and factual knowledge of art history.

We show an overview of the number of prompts per category in Table \ref{tab:full_dataset_distribution}.

\subsection{Collection of Reference Images} \label{subsec:references}
For all categories except Landscapes, we retrieve at least one reference image to support the annotation task by providing suitable visual examples. For Buildings and Animals, we use SPARQL queries to crawl suitable images for the subjects from Wikidata. For Artworks, we obtain at least ten examples per period from Google Arts \& Culture\footnote{\url{7https://artsandculture.google.com/}}. For Maps, we source examples from OpenHistoricalMap\footnote{\url{openhistoricalmaps.com}}, a freely usable platform maintained by a community of mappers and historians. Licensing and copyright considerations are discussed in the Appendix \ref{app:images_copyright}.

\subsection{Collection of Expected Values}

To provide additional guidance for annotators, we compile short textual descriptions of what a correct image might depict. We refer to these as ``expected values''. To obtain them, we prompt GPT 4 for adjectives, single words, and brief phrases that describe the appearance implied by each subject–temporal cue pair. For instance, expected values for an image of a 1-month-old mammal include \emph{tiny}, \emph{infant}, and \emph{juvenile}. We provide a list of all expected values in the Appendix \ref{app:exp_values}.

\section{Evaluating Temporal Knowledge in Multimodal Models} \label{sec:methodology}
We conduct a human evaluation on a subset of the TempViz dataset, yielding carefully curated ground-truth annotations that serve two purposes: (i) to rigorously assess the extent to which text-to-image models successfully encode and express temporal information during image synthesis, and (ii) to systematically evaluate the validity and reliability of automated evaluation measures in their capacity to judge the generated images accurately.

\subsection{Human Evaluation}
\paragraph{Task Description} For each instance, we present a human annotator with the input prompt and a model generation to assess. Additionally, we provide a reference image, and/or a list of expected values as discussed in \S\ref{subsec:references}. The annotator then needs to answer each of the following three questions (possible answers: \emph{yes}/ \emph{no}):

\begin{itemize}
\setlength{\itemsep}{0pt}
    \item[\textit{Q1}] \emph{Is the image free from any visual generation errors?}
    \item[\textit{Q2}] \emph{Is the subject present in the image?}
    \item[\textit{Q3}] \emph{Did the model apply the required temporal knowledge correctly?}
\end{itemize}
This way, we clearly separate concerns pertaining to the \emph{image quality (Q1)} from concerns that relate to the models' instruction following behavior. Moreover, here, we clearly distill general issues with depicting a \emph{subject (Q2)} from issues with following the \emph{temporal cues (Q3)}. 


\paragraph{Annotator Selection}
The annotation task was performed by two annotators from our institution who were (a) familiar with the concept of AI-based image generation, (b) proficient in the English language, and (c) familiarized with the task both via guidelines and direct training. Both annotators labeled the created images by the different models independently from one another. 

Despite employing only two annotators, we are confident in the robustness of our annotations for several reasons. First, many categories (e.g., \textit{Animals}, \textit{Landscapes}) exhibit low subjectivity and are less prone to annotator bias. Second, for more subjective categories (\textit{Artworks}, \textit{Buildings}), resource constraints prevented us from hiring domain experts, but we mitigated subjectivity through a tightly controlled annotation setup: we used on-site annotators rather than crowdsourcing, allowing direct communication and real-time clarification. We also provided reference images for subjective categories to align expectations and reduce noise. 

\subsection{Automatic Evaluation}
Given the cost and impracticality of large-scale human evaluation, we investigate several automated methods for reliably detecting temporal cues in images. Specifically, we employ two established image–text evaluation metrics (e.g., CLIPScore and Captioning), and query a suite of VLM-as-a-judge approaches using various strategies. In the following, we present all the approaches we compare.

\paragraph{Baseline} We compare all our results with a simple baseline computed by simply predicting the binary final label at random.  

\paragraph{CLIPScore} We adopt the established and widely used CLIPScore proposed by \citet{hessel-etal-2021-clipscore}. Therefore, we use a CLIP-based model to embed both the text prompt used to generate the image and the image itself, and then compute the cosine similarity between the corresponding embeddings to quantify prompt–image alignment.

\paragraph{Captioning} For the caption-based evaluation, we invert this procedure. We first employ a BLIP-based model to generate a descriptive caption for each generated image. We then use a sentence-embedding model to encode both the original prompt and the generated caption, and compute the cosine similarity between these embeddings.

For both CLIPScore and caption-based evaluation, we normalize the prompts prior to embedding by stripping imperative prefixes (e.g., ``Generate'') so that only the semantic content describing the intended depiction remains. Because both methods yield continuous similarity scores while the annotation labels are binary, we report the Point Biserial Correlation (PBS) between the similarity values and the ground-truth labels to enable a principled comparison.

\paragraph{Decompositional VQA} Following \citet{hu2023tifa}, we use an auxiliary LLM to decompose the prompt into fine-grained questions, enabling VQA models to assess nuanced prompt–image alignment. Given a text-to-image prompt, the LLM extracts salient elements and generates multiple-choice and open-ended questions with corresponding correct answers for each. Depending on the prompt complexity, the LLM generates a different number of question-answer pairs, resulting in at least seven questions per prompt. Subsequently, a VLM model is used to answer each question, given the T2I-generated image. Finally, if a certain threshold of questions was answered correctly, the image can be labelled as representing the prompt correctly. We provide an example in Appendix~\ref{app:autom_Eval}.

\paragraph{Direct VLM Judging}
Finally, we evaluate several variants of VLM-as-a-judge prompting: a VLM is asked whether the image correctly depicts the prompt and returns a judgment. We test multiple VLMs and compare strategies aimed at probing a VLM's capacity to assess temporal cues, including few-shot prompting and an elaborate judging prompt in the style of \citet{lee2024prometheusvisionvisionlanguagemodeljudge} that provides task guidance and requests structured criteria and a ranking. Further details and full prompt templates are provided in Appendix \ref{app:autom_Eval}.

\section{Experiments and Results}
\label{sec:exp}
\subsection{Experimental Setup}
\paragraph{Data} With our experiments, we address two questions: (1) whether T2I models are capable of incorporating temporal knowledge within their image generation process, and (2) how well current automatic evaluation approaches capture these fine-grained visual cues. To accurately answer both of these questions, we collect human annotations on a stratified sample of \textsc{TempViz} comprising 500 prompts with corresponding generations, balanced across categories and models.

\paragraph{Image Generation} We evaluate five T2I models on temporal knowledge, selected for public availability, architectural diversity, usage in prior research, and generation quality. Specifically, we compare \texttt{stable-diffusion-v1-5 (SDv1.5)} \cite{Rombach_2022_CVPR}, relying on the original Stable Diffusion architecture, with \texttt{sdxl-turbo (SDXL-T)}, \texttt{stable-diffusion-xl-base-1.0 (SDXL-B)}, and \texttt{stable-diffusion-3.5-large (SDv3.5)} relying on the SDXL architecture with a three times larger UNet and a combination of multiple text encoders compared to the original architecture \cite{podell2023sdxlimprovinglatentdiffusion}. This is to analyze performance differences that might occur due to scaling effects. SDXL-T constitutes the distilled version of the original SDXL model using Adversarial Diffusion Distillation. We chose this model to investigate potential drops in performance due to the distillation process. Finally, we add \texttt{FLUX.1-dev (FLUX)} \cite{flux2023} to represent a different publisher, and thus likely a different training corpus. Detailed information about the models and their architectures can be found in Appendix~\ref{app:exp_details}.

\paragraph{Automatic Evaluation} 
To compute embeddings for CLIPScore, we use \texttt{CLIP-ViT-bigG-14-laion2B-39B-b160k}, a CLIP-based model \cite{Radford2021LearningTV} fine-tuned on the LAION-5B corpus \cite{laion5b}. For automatic image caption generation, we employ the BLIP model \texttt{blip-image-captioning-base} \cite{li2022blipbootstrappinglanguageimagepretraining}. Then, to estimate semantic similarity between generated captions and prompts, we embed both with the \texttt{all-MiniLM-L6-v2} Sentence-BERT model \cite{reimers-gurevych-2019-sentence} and compute cosine similarities for the resulting representations. For the standard VQA analysis, we evaluate three models with strong reasoning performance: the open model \texttt{Qwen2.5-VL-32B-Instruct (Qwen2.5-VL)} \cite{bai2025qwen25vltechnicalreport} and the two proprietary models \texttt{gpt-4o-mini-2024-07-18 (GPT-4o mini)} and \texttt{gpt-5-2025-08-07 (GPT-5)} \cite{openai2024gpt4technicalreport}. Finally, for the advanced VQA approach, we use \texttt{Llama-3.3-70B-Instruct} \cite{grattafiori2024llama3herdmodels} to generate suitable question-answer pairs. Additionally, we use \texttt{Qwen2.5-VL-32B-Instruct} to predict the answer given the generated image.

\subsection{Results: Human Evaluation}
\paragraph{Inter-annotator Agreement}
We calculate Cohen's $\kappa$ \cite{cohensk} for each category and question individually and show the results in Table \ref{tab:cohensk}. 

\setlength{\tabcolsep}{7pt}
\begin{table}
\small
    \centering
    \begin{tabular}{c|ccc}
    \toprule
       \textbf{Category} & \textbf{Image Quality} & \textbf{Subject} & \textbf{Temporal} \\
       \midrule
       Landscape & 0.8160  & 1.0 & 0.9122 \\
       Animals & 0.5508  & 1.0 & 0.9189 \\
       Buildings & 0.2801  & 0.7626 & 1.0 \\
       Maps & 0.2412  & 0.9054 & 1.0 \\
       Artworks & 0.3080  & 1.0 & 0.9192 \\
       \midrule
       Average & 0.5016 & 0.8970 & 0.9502\\
       \bottomrule     
    \end{tabular}
    \caption{Inter Annotator Agreement of the Human Evaluation. We present Cohen's $k$ for each question individually and the average per category.}
    \label{tab:cohensk}
\end{table}

Overall, we observe an average inter-annotator agreement of 0.78 Cohen’s $\kappa$ across the three questions and categories. Encouragingly, for the categories most relevant to our study, i.e., whether the subject is present in the image and whether temporal knowledge was applied correctly, we achieve an average agreement of at least 0.76. The lowest agreement, 0.5 $\kappa$, is found for the question concerning image quality. A qualitative analysis revealed differing intuitions between the two annotators: while the first annotator considered only visibly distorted images as low quality, the second also classified non-photorealistic images as visual generation errors. In the final label aggregation, we adopted the more liberal criterion of the first annotator (\textbf{Q1}). Disagreements among the annotators for the other questions (\textbf{Q2}, \textbf{Q3}) were resolved through guided discussions with one of the authors.


\begin{figure}[h]
    \centering
    \begin{subfigure}[b]{0.45\textwidth}
        \centering
        \includegraphics[width=\textwidth]{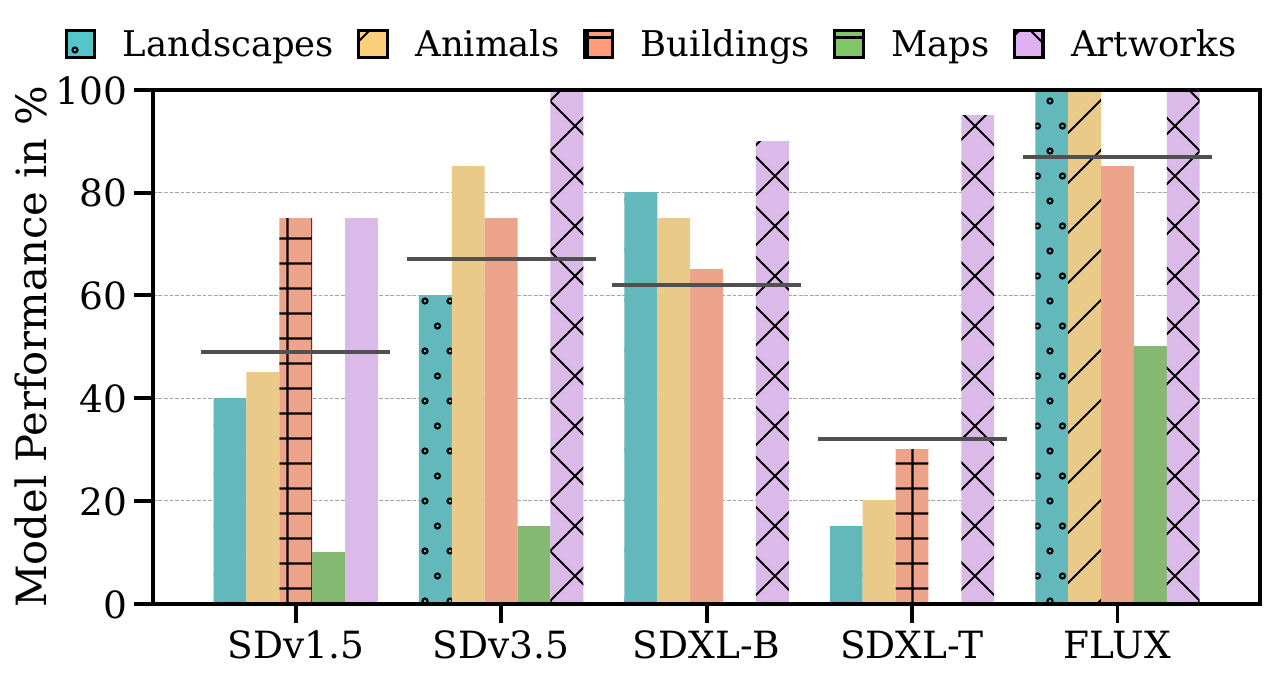}
        \caption{\textbf{Q1:} \% of images that do not contain any errors.}
        \label{fig:acc_q2_q3:quality}
    \end{subfigure}%
    \hfill
    \begin{subfigure}[b]{0.45\textwidth}
        \centering
        \includegraphics[width=\textwidth]{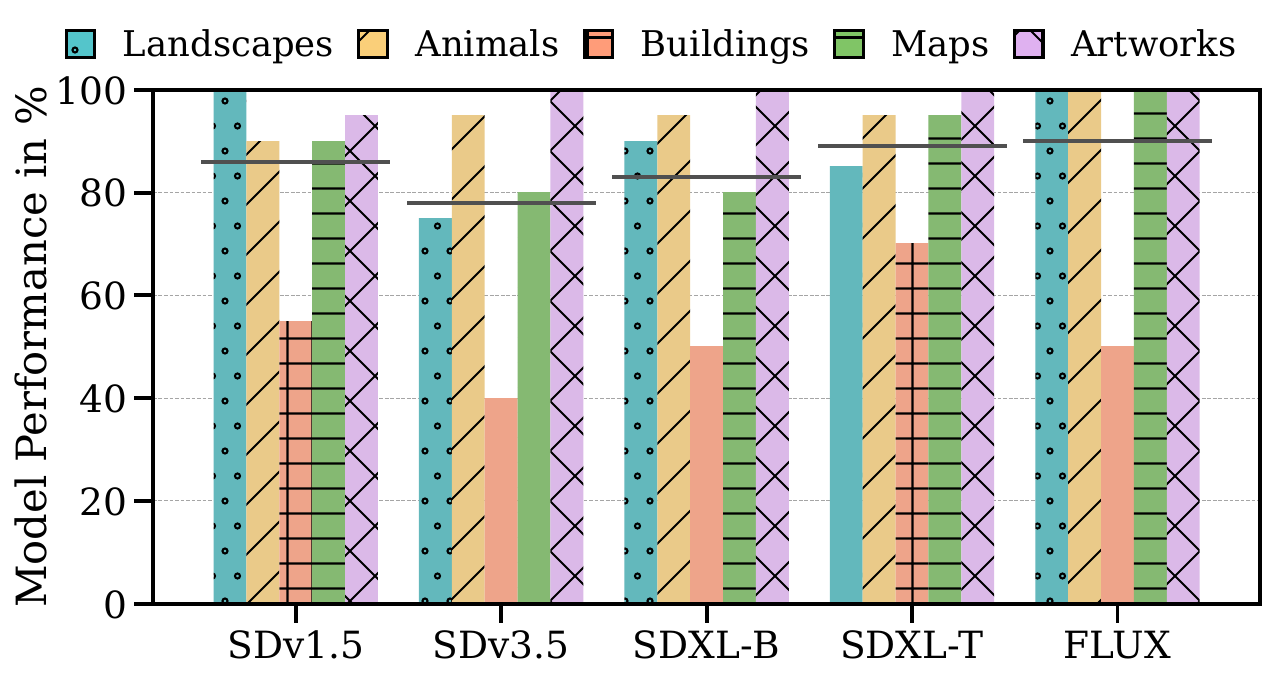}
        \caption{\textbf{Q2:} \% of images that present the correct subject.}
        \label{fig:acc_q2_q3:subject}
    \end{subfigure}%
    \hfill
    \begin{subfigure}[b]{0.45\textwidth}
        \centering
        \includegraphics[width=\textwidth]{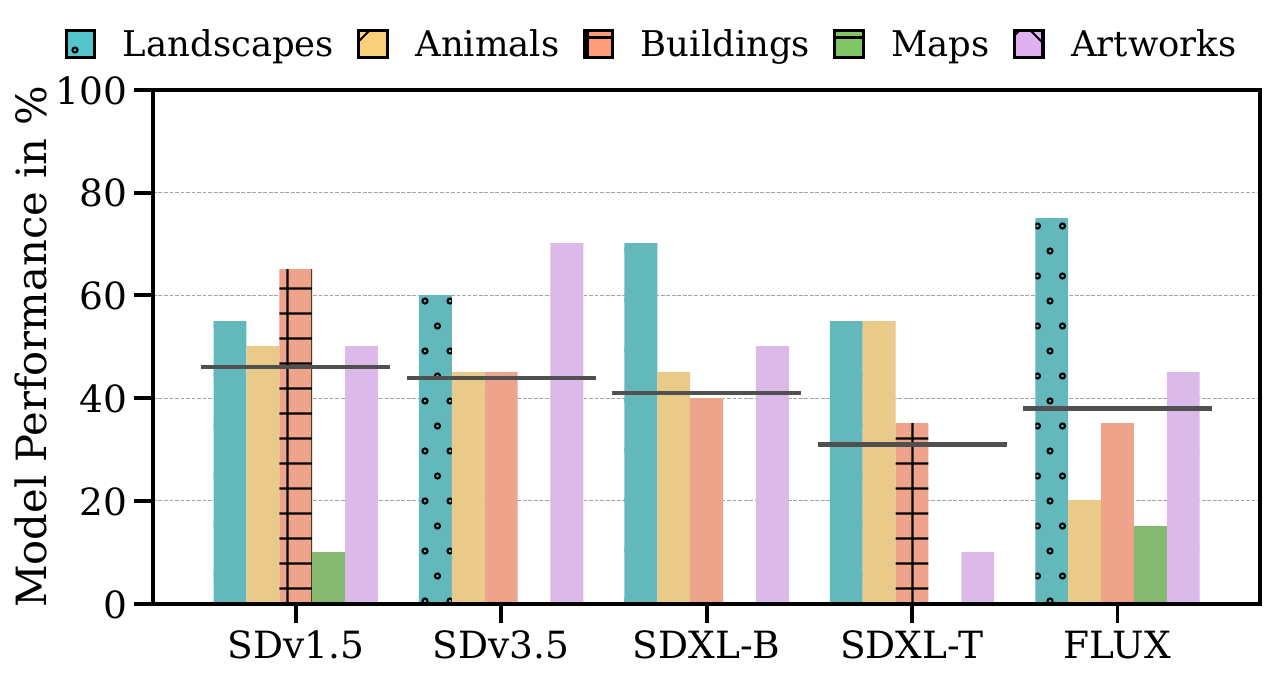}
        \caption{\textbf{Q3:} \% of images that correctly apply the required temporal knowledge.}
        \label{fig:acc_across_models:temporal}
    \end{subfigure}
    \caption{Accuracy per Category in \% of images that correctly represent the respective concept. We present the results for each tested T2I model separately. Grey bars indicate mean accuracy per model.}
    \label{fig:acc_across_models}
    \vspace{-1em}
\end{figure}

\paragraph{Temporal Knowledge in T2I models}
Utilizing the harmonized judgments obtained through our annotation study, we now discuss the T2I models' generations regarding their quality, subject presence, and application of temporal knowledge. 

\begin{figure*}[t]
    \centering
    \includegraphics[width=0.75\linewidth]{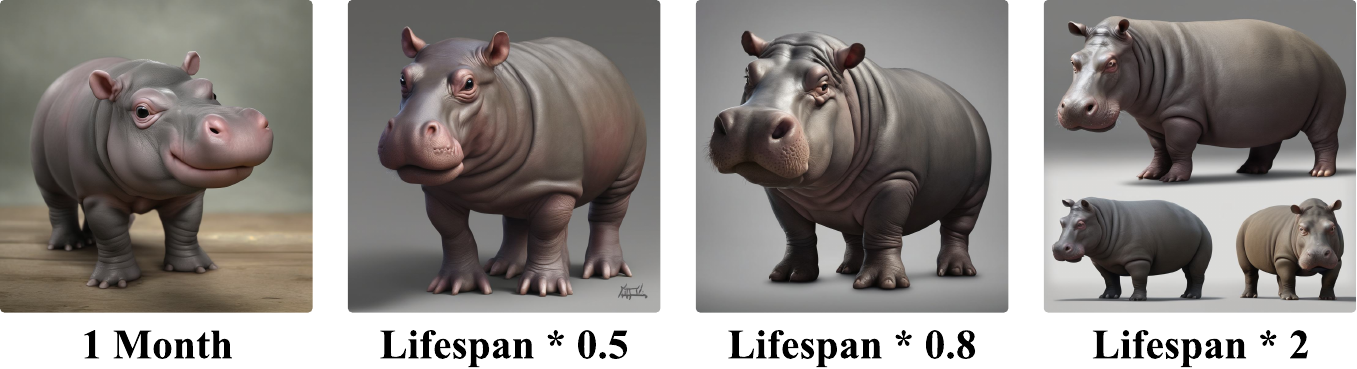}
    \caption{Example images generated by SDXL-B. We present an example from the Animals category for the subject \textit{Dwarf Hippo}, shown at different stages of its life cycle that serve as temporal cues.}
    \label{fig:happy_hippo}
    \vspace{-1em}
\end{figure*}

Figure~\ref{fig:acc_across_models} reports model accuracy on the three questions. For \textbf{Q1} (image quality), SDXL-T attains the lowest score, with only on average 32\% of images generated without visible generation errors, whereas FLUX achieves the highest average accuracy of 87\% across categories. The \emph{Maps} category emerges as the most challenging; even FLUX yields only 50\% images without visual errors, and both SDXL-T and SDXL-B fail to produce any coherent outputs in this setting. By contrast, the \emph{Artworks} category shows the largest share of images without apparent errors. This pattern could partially be explained by the fact that annotators could interpret generation errors within this category as stylistic variation within artistic imagery, which can mask underlying generation defects.

Turning to \textbf{Q2} (subject presence), we find that all models generally demonstrate high accuracy in correctly generating the requested subject (e.g., a horse) of most categories, but struggle for the \emph{Artworks}, where correct depiction of the subject is closely coupled with the application of temporal knowledge such as era and style requirements. This observation suggests that the basic object-level complexity of the prompts is not the primary performance bottleneck for the majority of categories.

Finally, in Figure \ref{fig:acc_across_models:temporal} we present the results for \textbf{Q3} (temporal knowledge). Performance is uniformly low across categories, indicating that models struggle to integrate temporal cues. Notably, FLUX, despite leading on \textbf{Q1} and \textbf{Q2}, attains the second temporal accuracy at 38\%, whereas SD v1.5, although producing less coherent images overall, achieves the highest average at 46\%. As expected, all models score lowest for the \emph{Maps} category, underscoring the difficulty of the task. Strikingly, even for supposedly simpler categories such as \emph{Landscapes} and \emph{Animals}, models do not exceed 75\% temporal accuracy. 
Overall, these findings indicate the following: (\emph{i}) General image quality performance is not an indicator of the temporal knowledge encoded in the models. (\emph{ii}) Different categories demand distinct types of temporal understanding, each posing varying levels of difficulty for text-to-image (T2I) models -- but seemingly easier categories are not always easiest for all models. (\emph{iii}) Finally, current T2I models remain limited in their ability to encode and apply temporal knowledge in a consistent and systematic manner -- even when succeeding in subject placement.

In Figure \ref{fig:happy_hippo}, we show an example from SDXL-B for the subject \textit{Dwarf Hippo} across four temporal cues. In this example, the model largely captures age progression across the typical lifespan, and presumably even tries to depict a deceased or extinct-looking specimen in the right image.

\subsection{Results: Automatic Evaluation}
Next, we investigate the ability of automatic methods to judge the correct use of temporal cues specified in the prompt. As ground truth, we use the labels from our human annotation study. We compare traditional similarity metrics with vision–language judges under multiple prompting strategies.

\paragraph{Traditional Approaches}
Table \ref{tab:pbs_results} shows the results for the two traditional approaches: ClipScore and the caption-based evaluation. Overall, both methods are largely ineffective at capturing the fine-grained temporal cues targeted in our study. CLIPScore shows no significant correlation with any of the annotated labels. We hypothesize that these results are largely caused by CLIP's limited sensitivity to temporal expressions and their subtle visual consequences. This is consistent with prior work showing that metrics such as CLIPScore struggle with fine-grained or nuanced visual attributes \cite{bugliarello-etal-2023-measuring, wiles2025revisitingtexttoimageevaluationgecko}. Moreover, our use of CLIPScore compares each generated image to a single reference text, a setting that is less informative than CLIP's original retrieval-style use case in which multiple candidate captions are ranked relative to one another, which likely also contributes to the weak association even for subject presence. The one notable exception is CLIPScore's relatively better performance on the \textit{Artworks} category, which we hypothesize stems from artworks being particularly well-represented in CLIP's training data and from temporal cues being more tightly coupled with subject identity in this domain (e.g., ``Renaissance artworks'').

In contrast to the CLIPScore, prompt–caption similarity yields significant correlations, but these gains primarily reflect subject presence (\textbf{Q2}), with the strongest effects in Artworks and Buildings. In contrast, correlations for the temporal cues (\textbf{Q3}) are negative, indicating that caption-based measures can mislead when correctness depends on temporal cues rather than object identity.

\setlength{\tabcolsep}{8.5pt}
\begin{table}[]
\small
    \centering
    \begin{tabular}{llcr}
    \toprule
    \textbf{Approach} & \textbf{Dimension} & \textbf{Question} & \textbf{PBS} \\
    \midrule
     & Subject & \textbf{Q2} & 0.0158\\
ClipScore & Temporal & \textbf{Q3}  & -0.0329\\
 & Subj+Temp & \textbf{Q2+Q3} & -0.0394\\
 \midrule
     & Subject & \textbf{Q2} & 0.1571* \\
Captioning & Temporal & \textbf{Q3} & -0.1841*\\
 & Subj+Temp & \textbf{Q2+Q3} & -0.0955*\\
\bottomrule
    \end{tabular}
    \caption{Results for traditional evaluation approaches. We present point biserial correlations with human annotations for \textbf{Q2} and \textbf{Q3}, plus their combined judgment. Asterisks indicate statistical significance.}
    \label{tab:pbs_results}
    \vspace{-1em}
\end{table}

\begin{figure}
    \centering
    \includegraphics[width=\linewidth]{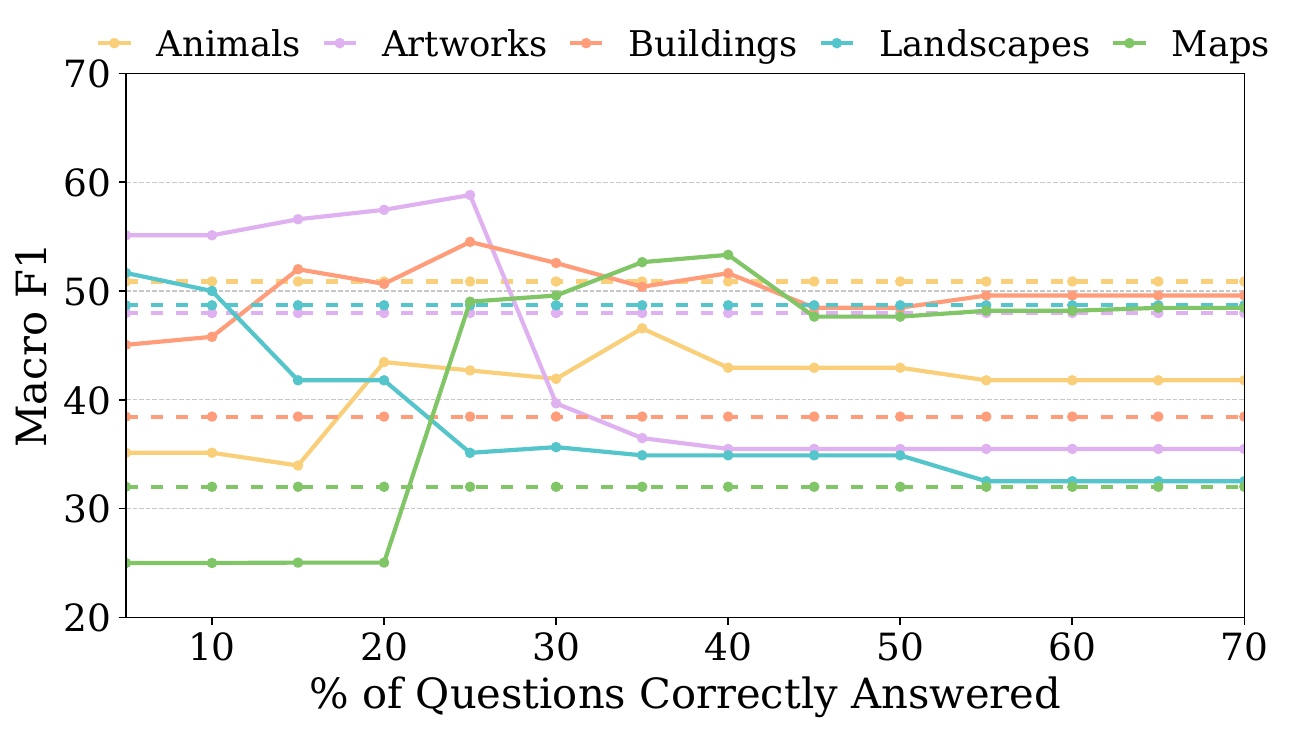}
    \caption{Decompositional VQA results. We report macro F1 scores (y-axis) across categories for varying answer correctness thresholds (y-axis). Each threshold indicates the minimum percentage of correctly answered questions required to classify an image as successfully reflecting the temporal prompt.
    Dashed lines indicate the random baseline for each category.}
    \label{fig:advanced_vqa_results}
\end{figure}

\paragraph{Decompositional VQA} 
Figure \ref{fig:advanced_vqa_results} shows the macro-F1 scores of decompositional VQA across different thresholds for correctly answered LLM-generated questions. The method is evaluated against the combined human judgment for \textbf{Q2} and \textbf{Q3}, assessing whether the full prompt is accurately represented in the image. For \textit{Animals}, the method fails to surpass the baseline at any decision threshold. Across the remaining categories, macro-F1 never exceeds 60\%, and no single question-accuracy threshold yields consistently strong performance. Taken together, these patterns indicate that the approach is not yet adequate for reliable automatic assessment of temporal cues in images.

\setlength{\tabcolsep}{5.5pt}
\begin{table}[t]
    \centering
    \small
\begin{tabular}{lp{1cm}lccc}
\toprule
    \textbf{Model} &  \textbf{Strategy} & \textbf{Prompt} & \textbf{P} &  \textbf{R} & \textbf{F1} \\
\midrule
 \multirow{1}{*}{Baseline} &&  & 48.4 & 48.3 & 47.7\\
\midrule
\multirow{5}{*}{Qwen-VL} & \multirow{2}{*}{0-Shot} & Simple &  64.3 &  60.1 & 60.0 \\
 &  & Instr. &      64.8 & 64.4 &   64.6\\
\cmidrule(lr){2-6}
 & \multirow{2}{*}{3-Shot}  & Simple &     61.9 & 62.7 &   62.0\\
 &  & Instr. & \underline{65.9} &  \underline{67.2} & \underline{65.8}\\
  \midrule
\multirow{5}{*}{GPT-4o mini} & \multirow{2}{*}{0-Shot} & Simple & \underline{64.4} &  61.2 & 61.5 \\
 & & Instr. & 61.0 &  59.5 & 59.8 \\
\cmidrule(l){2-6}
 & \multirow{2}{*}{3-Shot} & Simple &  62.7 &  \underline{61.7} & \underline{62.0} \\
 &  & Instr. & 59.3 &  58.0 & 58.1 \\
  \midrule
\multirow{5}{*}{GPT-5} & \multirow{2}{*}{0-Shot} & Simple & 73.2 &  66.0 & 66.8 \\
 & & Instr. & 75.5 &  68.3 & 69.4 \\
\cmidrule(l){2-6}
 & \multirow{2}{*}{3-Shot} & Simple & \textbf{75.7} &  \textbf{70.9} & \textbf{72.0} \\
 &  & Instr. & 74.6 &  69.5 & 70.5 \\
\bottomrule
\end{tabular}
    \caption{Direct VLM Judging: We present macro Precision (P), Recall (R), and F1 against human annotations for the combined judgments of \textbf{Q2} and \textbf{Q3} (full prompt).}
    \label{tab:results_prompt_vqa}
\end{table}

\paragraph{Direct VLM Judging}
We report aggregated results for the direct VLM judging approach in Table \ref{tab:results_prompt_vqa}. Finer-grained results are provided in Appendix~\ref{app:exp_details}. We evaluate two prompting styles: Simple, which asks whether the image correctly depicts the prompt, and Instructed, which provides task guidance and requests a rating. For each style, we consider zero-shot and three-shot variants.

Encouragingly, all three models exceed the random baseline across all prompting strategies. GPT-5 attains the strongest overall performance but peaks at a macro F1 of 72\%, well below reliable accuracy. For both GPT-5 and GPT-4o mini, the \textit{Simple} prompts outperform the \textit{Instructed} prompts, indicating that while the models grasp the task without guidance, the relevant temporal cues might be too subtle and remain difficult to assess consistently. Surprisingly, Qwen-VL even outperforms GPT-4o mini with the \textit{Instructed} prompt. Moreover, few-shot prompting with out-of-distribution exemplars yields gains for all models, with the largest increase for GPT-5. Overall, even the most recent proprietary system falls well short of robust performance, indicating that the VQA-as-judge paradigm is not yet sufficient for dependable automatic evaluation of temporal knowledge in generated images.

\begin{figure}
    \centering
    \includegraphics[width=0.9\linewidth]{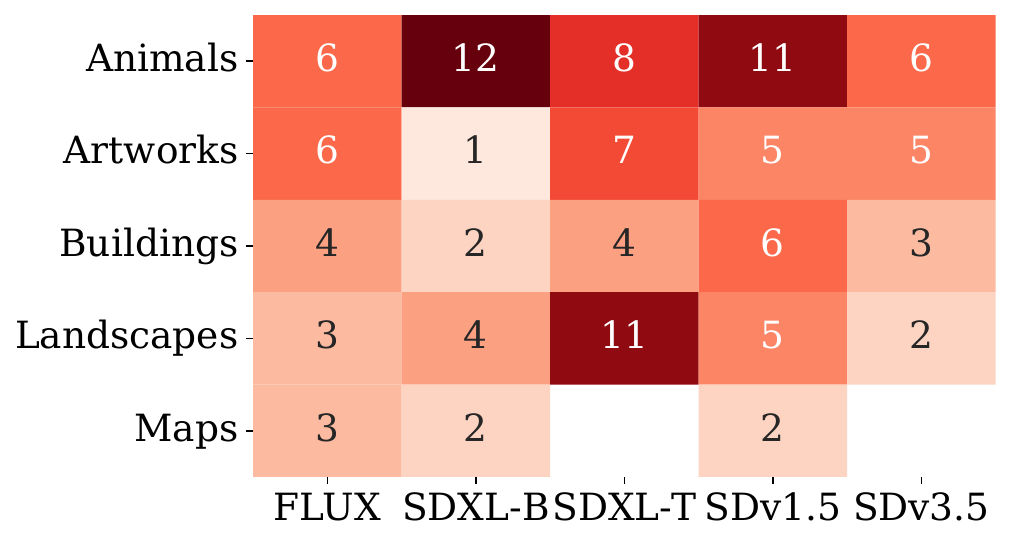}
    \caption{GPT-5 error distribution across categories and T2I models for the best performing prompting strategy.}
    \label{fig:error_dist}
    \vspace{-1em}
\end{figure}

\paragraph{Error Analysis}
Taking a closer look, Figure \ref{fig:error_dist} shows GPT-5's failure distribution to examine systematic patterns. Despite human annotations identifying \textit{Buildings} and \textit{Maps} as the most challenging categories, GPT-5 most frequently misjudges the presumably simpler \textit{Animals} category across T2I models. Conversely, \textit{Maps} exhibits the fewest errors, suggesting that its temporal cues (or their absence) are easier for the model to detect. Apart from an outlier in \textit{Landscapes} for SDXL-T, which aligns with the low image quality noted in human evaluation, errors are otherwise relatively evenly distributed. Overall, GPT-5's evaluation reliability is not strictly correlated with human-perceived task difficulty. Instead, the model appears more sensitive to visually ambiguous or less distinctive temporal cues (e.g., aging in animals) than to structurally constrained ones (e.g., historical maps).

\section{Conclusion}
We introduced \textsc{TempViz}, the first dataset specifically designed to evaluate the temporal knowledge in T2I across a diverse set of temporal dimensions. Using \textsc{TempViz}, we studied the ability of five T2I models to incorporate temporal knowledge into image generation. To this end, we conduct a human evaluation and additionally compare the obtained judgments against several automated judging schemes. We find that all T2I models we tested show only low to moderate capabilities in applying the required knowledge across the different temporal categories of \textsc{TempViz}. We further find that automatic evaluation approaches fail to reliably judge the temporal cues depicted in images and that their mistakes do not necessarily align with those of humans. In sum, \textbf{there is thus a pressing need for both improving the understanding as well as the generation of temporal cues in images}. We hope that our findings and our novel \textsc{TempViz} resource can motivate further research in this direction.

\section*{Limitations}
Our work has several limitations that need to be acknowledged. First, due to resource constraints, our work is limited to five T2I models. However, we have decided to use frequently deployed models that exhibit state-of-the-art performance and hope that our benchmark will also be used to evaluate further T2I models. Second, for the same reason, we relied on only two annotators to evaluate the generated images, which may introduce bias or limit the robustness of the results. Third, while our evaluation approach offers a starting point, more sophisticated and varied evaluation metrics could provide a deeper understanding of the models' capabilities in integrating temporal knowledge in the image generation process. Fourth, while the reference images we provide aid dataset annotation, they serve only as a silver standard and cannot provide sufficient coverage for reference-based evaluation. Landscape prompts such as ``A lake in Europe'' are inherently too diverse to capture with single reference images. For Animals, we typically have only one image per species, lacking the multiple aging stages required to match our temporal prompts. Similarly, Buildings often have reference images for only one temporal state (e.g., the historical structure before deconstruction) without corresponding images of later states. Finally, certain categories in our dataset require specific domain knowledge to judge accurately, which could impact the reliability of those evaluations. However, by providing reference images for more difficult categories, we are confident that we can minimise these effects.

\section*{Acknowledgements}
The work of Carolin Holtermann and Anne Lauscher is funded by the Excellence Strategy of the German Federal Government and the Federal States.

\bibliography{custom}

\clearpage

\appendix
\label{sec:appendix}

\twocolumn

\section{TempViz Creation}

\subsection{Prompt Formulation Variations}

\begin{table}[h]
    \centering
    \small
    \begin{tabular}{c p{6cm}}
    \toprule
     \textbf{ID} & \textbf{Prompt Template} \\
     \midrule
     1  &  \textit{Give me a photorealistic image of \{subject\} in \{temporal cue\}.}\\
     2  &  \textit{Produce a photorealistic image of \{subject\} in \{temporal cue\}.}\\
     3  &  \textit{Generate a photorealistic image of \{subject\} in \{temporal cue\}.}\\
     4  &  \textit{Create a photorealistic image of \{subject\} in \{temporal cue\}.}\\
     \bottomrule
    \end{tabular}
    \caption{Generic prompt templates used for each subject $s_i$ and temporal cue $t_i$ combination $s_i\times t_i$ in \textsc{TempViz}.}
\end{table}

\subsection{Distribution Maps Category}\label{app:maps_distr}
\begin{figure}[h]
    \centering
    \includegraphics[width=\linewidth]{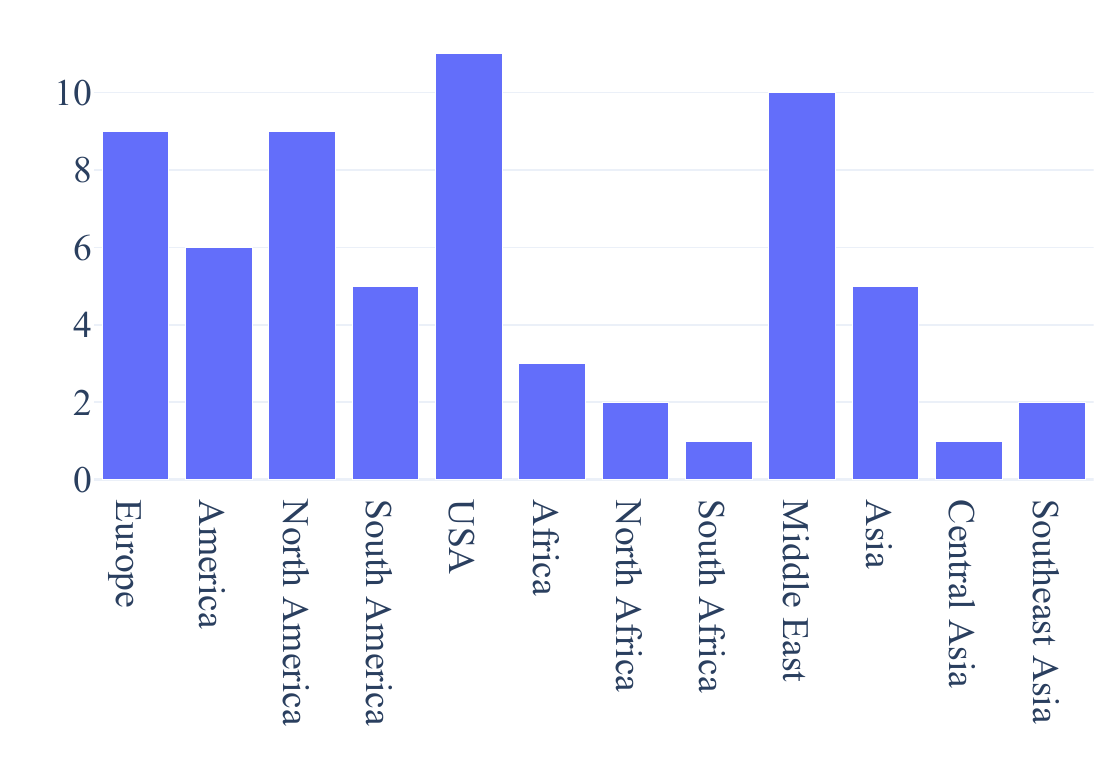}
    \caption{Number of suggestions per region in the Maps category of TempViz}
    \label{fig:maps_distribution}
\end{figure}

Figure \ref{fig:maps_distribution} shows the distribution of regions we use to prompt the Maps category of TempViz.

\clearpage
\onecolumn
\subsection{Reference Images}\label{app:images_copyright}

\begin{table*}[h]
    \centering
    \begin{tabular}{c|p{6cm}|p{6cm}}
    \toprule
       Category  & Source &  License \\
       \midrule
        Animals,  & Wikidata   & Creative Commons C0 \\
        Buildings& \url{https://www.wikidata.org/wiki/Wikidata:Licensing}& \url{https://creativecommons.org/publicdomain/zero/1.0/} \\
        \hline
        Maps & Open History Map  & Creative Commons BY 4.0 \\
        &\url{https://map.openhistorymap.org/} & \url{https://creativecommons.org/licenses/by/4.0/} \\
        \hline
        Artworks & Google Arts and Culture \url{https://artsandculture.google.com/} & The licenses under which Google Arts and Culture makes the images available are very opaque, so we have decided to make these images available in our dataset only through links to the corresponding website.  \\
        \bottomrule
    \end{tabular}
    \caption{A list of the licenses under which the images in our dataset were published.}
    \label{tab:licenses}
\end{table*}

\clearpage

\subsection{Expected Values}\label{app:exp_values}

\begin{table}[h]
\small
    \centering
    \begin{tabular}{lp{0.7\textwidth}}
        \toprule
        \textbf{Art period} & \textbf{Expected Values} \\
        \midrule
        Ancient Art & symbolic, preserved, symbolism, craftsmanship, ritual, early civilization, religious and methodological themes, symbolic imagery    \\
        Baroque Art & dramatic, ornate, grandiose, lavish, opulent, vivid, ornamentation, grandeur, lavishness, opulence, vibrancy, grandeur in scale and detail, elaborate ornamentation, complex composition   \\
        Contemporary Art & abstract, experimental, minimalist, abstraction, minimalism, diverse forms, diverse cultural perspectives, experimental techniques  \\
        Impressionism & luminous, expressive, vibrant, fluid, evocative, luminosity, expression, vibrancy, fluidity, effects of light and color, impression of a scene, scenes of everyday life, landscapes, urban settings, light and weather, less detail   \\
        Medieval Art & religious, ornamental, gothic, byzantine, devotion, symbolism, ornamentation, monumentality, strong religious influence, ecclesiastical purpose, christian themes, romanesque, spiritual aesthetic, two-dimensional aesthetic \\
        Modern Art & religious, ornamental, gothic, byzantine, devotion, symbolism, ornamentation, monumentality, strong religious influence, ecclesiastical purpose, christian themes, romanesque, spiritual aesthetic, two-dimensional aesthetic \\
        Neoclassicism & formal, realistic, monumental, sculptural, formality, realism, monumentality, sculpturality, clarity, elegant simplicity, formal composition, restrained emotional expression, ideals of harmony, proportion and disciplined technique, enlightenment \\
        Post-Impressionism & vibrant, expressive, symbolic, structured, geometric, bold, vibrancy, symbolism, geometry, boldness, focus on light and color, emphasis on deeper symbolic content, structural forms, vivid and emotive use of line and color, emotional depth, structured compositions, expressive, sometimes geometric forms \\
        Realism & detailed, naturalistic, mundane, detail, authenticity, observation, naturalism, depicting everyday scenes, depicting people in a true-to-life manner, meticulous detail, naturalistic depiction of subjects, still lifes \\
        Renaissance Art & harmonious, realistic, detailed, classical, proportional, luminous, harmony, detail, realistic representation, precise anatomy, linear perspective, depth \\
        Rococo & ornamental, lavish, fluid, delicate, sensuous, lavishness, delicacy, elaborate ornamentation, intricate detailing, elegant and decorative aesthetics, lightness, curvaceous forms, themes of love, themes of nature, soft colors, asymmetrical designs, sculptural elements \\
        Romanticism & emotive, sublime, passionate, mystical, exotic, sublimity, passion, mysticism, exoticism, sublime beauty of nature\\ 
        \bottomrule
    \end{tabular}
    \caption{Expected values for each art periods in the Artworks prompt category}
    \label{tab:Appendix EV_artworks}
\end{table}

\begin{table}[h]
\small
    \centering
    \begin{tabular}{lp{0.7\textwidth}}
        \toprule
        \textbf{Prompted Lifespan} & \textbf{Expected Values} \\
        \midrule
        1 month & tiny, soft, infant, juvenile, suckling, offspring, baby, small, bright-eyed, smooth, bright eyes, wide and open eyes, cuddly and endearing presence   \\
        0.5*lifespan & mature, vigorous, strong, healthy, adult, muscular, toned, fit, agile, vibrant, sturdy, strong and healthy, muscular and toned body, vitality and good health, vibrant and sturdy appearance, prime of their life   \\
        0.8*lifespan & elderly, aged, grey, wrinkled, signs of aging, graying fur, thinning fur, weathered appearance, wrinkled skin   \\
        2*lifespan & lifeless, deceased, dead, late, passed, perished, death, passing, decease, mortality   \\
        \bottomrule
    \end{tabular}
    \caption{Expected values for each lifespan in the Animals prompt category}
    \label{tab:Appendix EV_animals}
\end{table}

\begin{table}[h]
\small
    \centering
    \begin{tabular}{lp{0.7\textwidth}}
        \toprule
        \textbf{Prompted Lifespan} & \textbf{Expected Values} \\
        \midrule
        demolished & ruined, dilapidated, shattered, decayed, wrecked, ravaged, devastated, demolished, destroyed, torn off, rubble, debris, ruins, wreckage, fragments, remains, state of ruin   \\
        \bottomrule
    \end{tabular}
    \caption{Expected values for the current state \textit{"demolished"} in the Buildings prompt category}
    \label{tab:Appendix EV_Buildings}
\end{table}

\begin{table}[h]
\small
    \centering
    \begin{tabular}{p{0.23\textwidth}p{0.7\textwidth}}
        \toprule
        \textbf{Season} & \textbf{Expected Values} \\
        \midrule
        Winter, december, january, february & frosty, snowy, icy, snow, ice, frost, frosty and icy \\
        Spring, march, april, may & blooming, lush, blossoming, green, greenery, flowers, bloom, fresh greenery, trees and plants covered in buds and blossoms, nature's renewal \\
        Summer, june, july, august & sunny, bright, green, warm, sunshine, greenery, warmth, blue skies, bright sunshine\\
        Fall, september, october, november & golden leaves, colored leaves, orange leaves, warm-hued landscape \\
        \bottomrule
    \end{tabular}
    \caption{Expected values for the different seasons in the Landscape prompt category}
    \label{tab:Appendix EV_Time}
\end{table}

\clearpage
\section{Experimental Details}\label{app:exp_details}
Table \ref{tab:models} shows all T2I models we analyze and details on their architecture. All models that we use for our experiments are loaded using the Hugging Face Transformers library\footnote{\url{https://huggingface.co/docs/transformers}} and we run all experiments on 4 NVIDIA RTX A6000 GPUs.

\begin{table*}[h]
    \centering
    \small
    \begin{tabular}{l|l|c|c}
    \toprule
      Modelname &   Text Encoder & Diffuser \\
      \midrule
        FLUX Dev & CLIP-G/14, CLIP-L/14, T5 XXL & MM-DiT \\

        Stable Diffusion XL - Base  & CLIP ViT-L/14, OpenCLIP ViT-bigG/14 & Latent Diffusion U-Net\\
        Stable Diffusion XL - Turbo  & CLIP ViT-L/14, OpenCLIP ViT-bigG/14 & Latent Diffusion U-Net (distilled, ADD)\\
        Stable Diffusion v1.5  & CLIP ViT-L/14 &  Latent Diffusion U-Net \\
        Stable Diffusion v3.5  & CLIP G/14, CLIP L/14, T5 XXL & MMDiT (Diffusion Transformer)\\
        \bottomrule
    \end{tabular}
    \caption{Architectures of T2I models.}
    \label{tab:models}
\end{table*}

\clearpage

\twocolumn
\section{Annotator Disagreement}\label{app:annotator_disagreement}

\begin{figure}[h]
    \centering
    \begin{subfigure}[b]{0.45\textwidth}
        \centering
        \includegraphics[width=\textwidth]{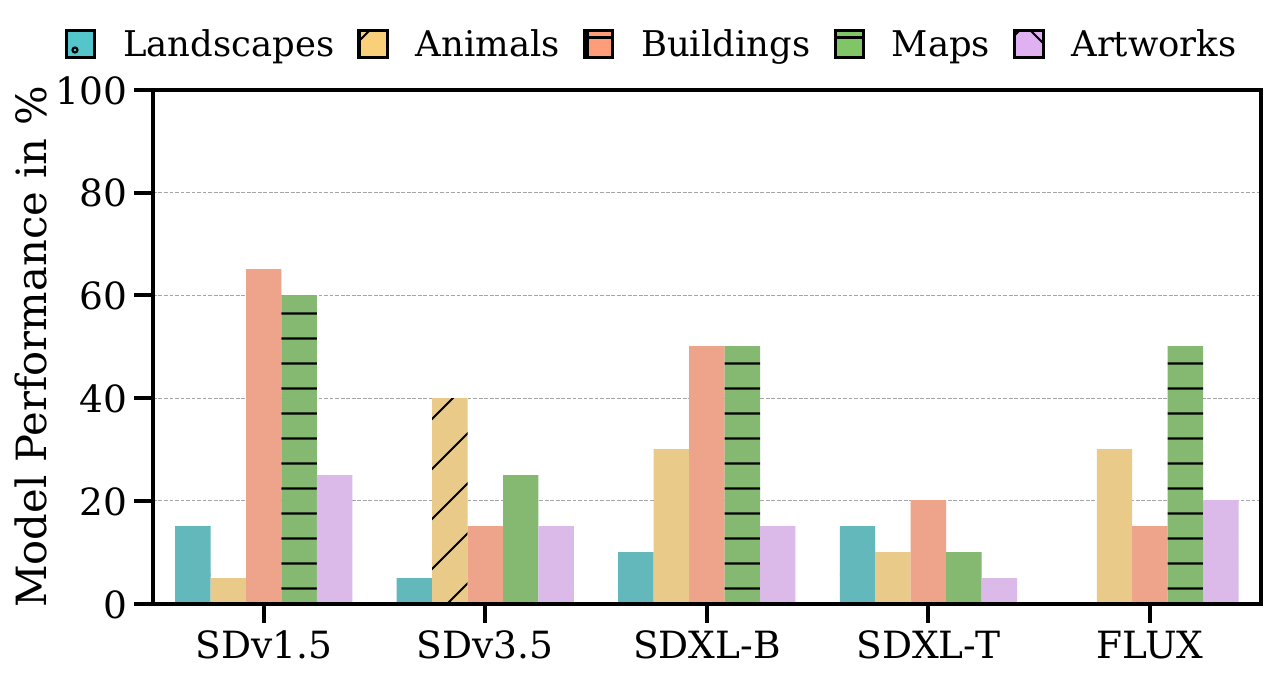}
        \caption{\textbf{Q1:} Disagreement in \% for image quality errors.}
    \end{subfigure}%
    \hfill
    \begin{subfigure}[b]{0.45\textwidth}
        \centering
        \includegraphics[width=\textwidth]{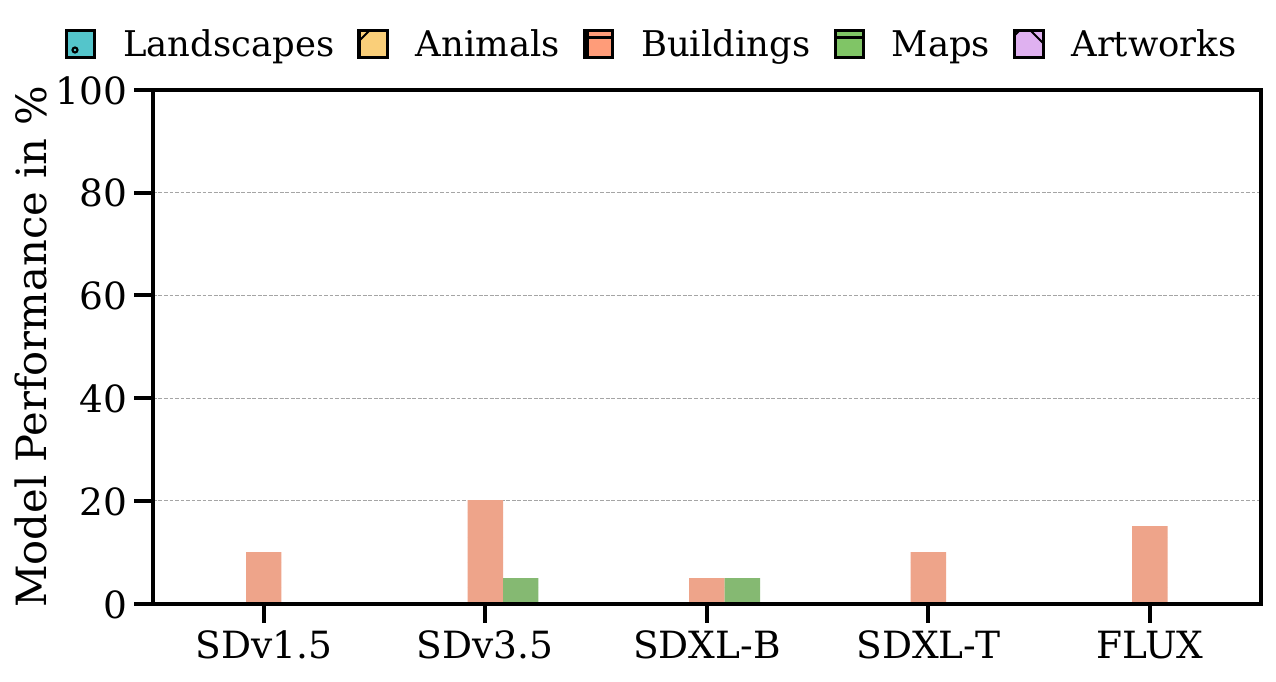}
        \caption{\textbf{Q2:} Disagreement in \% for correct subject.}
    \end{subfigure}%
    \hfill
    \begin{subfigure}[b]{0.45\textwidth}
        \centering
        \includegraphics[width=\textwidth]{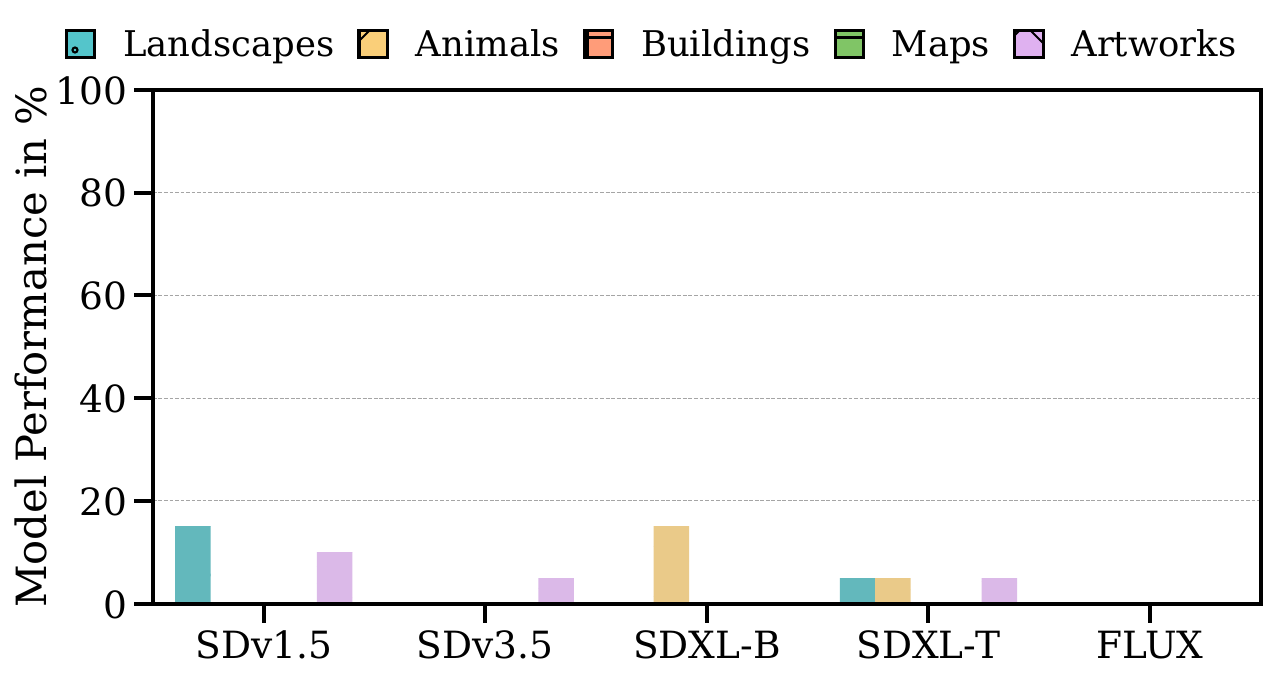}
        \caption{\textbf{Q3:} Disagreement in \% for temporal knowledge.}
    \end{subfigure}
    \caption{Annotator Disagreement in \% of images for the respective annotation question.}
    \label{fig:anno_dis}
\end{figure}

\clearpage
\section{Automatic Evaluation}\label{app:autom_Eval}
\subsection{Prompting Strategies}

\paragraph{Decompositional VQA} We use the following prompt to generate suitable question-image pairs from each of the original T2I input prompts.

\begin{tcolorbox}[colframe=green!50!black, colback=green!5!white, 
                  title=LLM Prompt, fonttitle=\bfseries]
Given an image description, generate multiple-choice questions that can verify if the image description is correct for an image.
First, extract elements from the image description.
Then classify each element into a category (object, human, animal, food, activity, attribute, counting, color, material, spatial, location, shape, other). 
Finally, generate different questions for each element, open-ended and multiple choice and also the correct answer. For example: 'What is the main subject of the image? Answer: Pony'
Description: '\{\}'
    
\end{tcolorbox}

\paragraph{Example Questions by Llama3.3 70B}
In the following, we present example questions for the prompt \textit{``Create a photorealistic image of an Aardvark that is one month old.''} generated by Llama3.3 70B.

\begin{tcolorbox}[colframe=orange!50!white, colback=orange!5!white, 
                  title=Example Questions by the LLM, fonttitle=\bfseries]
[...]
\#\#\# Questions for Each Element:

\#\#\#\# For Aardvark (Animal):

1. **Open-ended question:** What type of animal is the main subject of the image?
   - **Answer:** Aardvark

2. **Multiple-choice question:** What is the main animal in the image?
   - **A)** Lion
   - **B)** Elephant
   - **C)** Aardvark
   - **D)** Giraffe
   - **Correct answer:** C) Aardvark

\#\#\#\# For One month old (Attribute - Age):

1. **Open-ended question:** How old is the Aardvark in the image?
   - **Answer:** One month old

2. **Multiple-choice question:** What is the age of the Aardvark depicted in the image?
   - **A)** One week old
   - **B)** One month old
   - **C)** Three months old
   - **D)** Six months old
   - **Correct answer:** B) One month old

[...]
\end{tcolorbox}

\paragraph{Direct VLM Judging}
We present the two prompting strategies that we use to guide the VLMs to analyze the generated image with respect to the input prompt.

\begin{tcolorbox}[colframe=blue!50!black, colback=blue!5!white, 
                  title=Standard Prompt, fonttitle=\bfseries]
Is the image correct for the following prompt:  \{\} Is the correct temporal knowledge applied (i.e., <temporal cue>)? Just reply with 'yes' or 'no'
    
\end{tcolorbox}

\begin{plainbox}[Judging Prompt]
As a professional ``Text-to-Image'' quality assessor, your task is to judge the performance of a
text-image model w.r.t. a certain criterion by evaluating the image generated from a specific prompt.
The criteria for evaluation are as follows:\
Rubrics:
'Assess how accurately the image reflects the given prompt. Check if the object attributes along with all temporal elements—such as signs of aging in animals, 
historical map borders, seasonal indicators in landscapes, lighting and shadows for time of day, architectural features that change over time, and stylistic 
traits of artworks from different epochs—are correctly represented.'
Please analyze step by step and provide the RATING using
the following scale: [``Not accept'', ``Accept'']. In this scale,
``Not accept'' represents images that do not accurately reflect the prompt.``Accept'' represents images that do accurately reflect the prompt, along and also the temporal elements.
Please do not generate any other opening, closing, or explanations. The output of the analysis and
rating should strictly adhere to the following format:
ANALYSIS: Provide your analysis here
RATING: Only provide your rating here.
Now, proceed with evaluating the image based on the prompt: '\{\}'
\end{plainbox}

\clearpage
\onecolumn
\subsection{Additional silver-standard T2I Evaluation Results}
We provide results on an additional set of 500 images for the combined questions Q2 + Q3 (that is, whether an image matches the prompt completely). This subset was labeled by two team members and one crowdsourced annotator to create a potential training set and was not included in the main paper because the annotations were less stringent and were collected only at the full prompt level. Given the limited size of the evaluation set, we release this additional subset as a silver standard benchmark to complement the primary annotations and further support our findings.

\begin{figure}[h]
    \centering
    \includegraphics[width=0.7\linewidth]{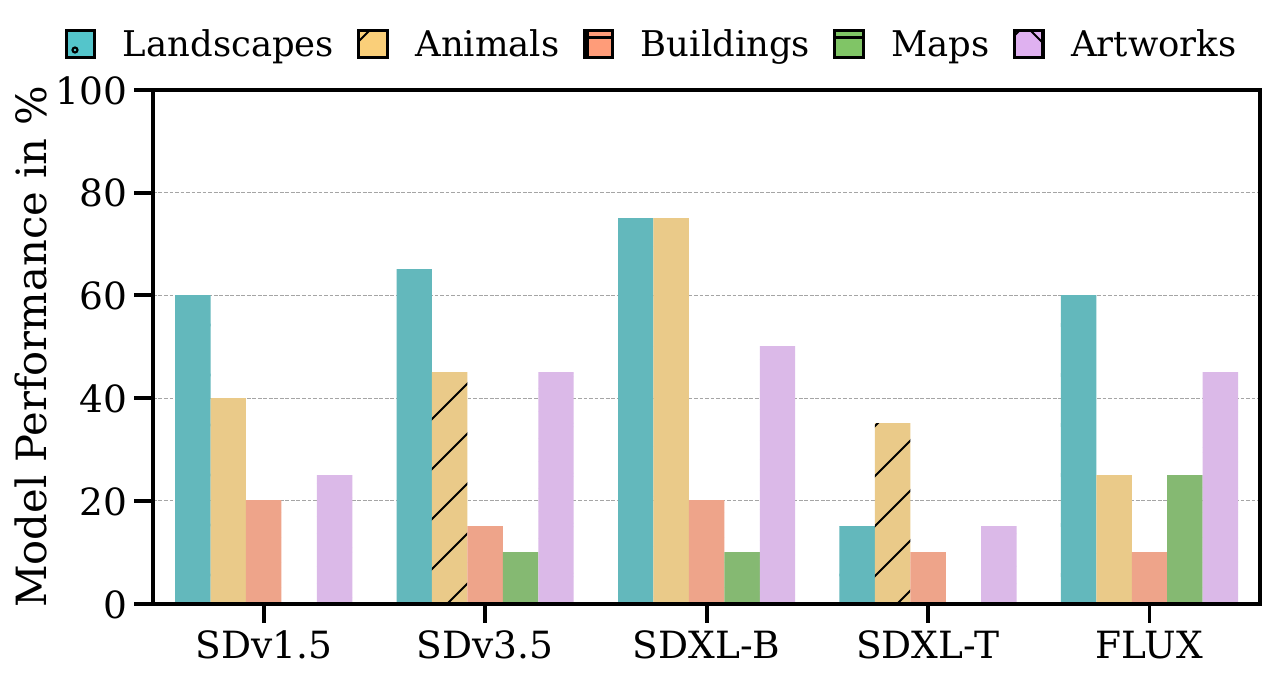}
    \caption{Accuracy per Category in \% of images that correctly represent the whole prompt. We present the results for each tested T2I model separately.}
\end{figure}

\clearpage
\onecolumn
\subsection{Additional Results for the Direct VLM Judging Approach}

\begin{table}[h]
    \centering
\begin{tabular}{llllccc}
\toprule
    \textbf{Model} &  \textbf{Strategy} & \textbf{Prompt} & \textbf{Category} & \textbf{P} &  \textbf{R} & \textbf{F1} \\
\midrule
 \multirow{1}{*}{Baseline} && & Overall & 48.4 & 48.3 & 47.7\\
\midrule
\multirow{20}{*}{Qwen-VL} & \multirow{10}{*}{0-Shot} & Simple   & Landscapes & 63.2 & 56.8 & 44.0 \\ 
 & & Simple & Animals & 79.3 & 51.2 & 39.3 \\ 
 & & Simple & Artworks & 74.3 & 72.6 & 70.8 \\ 
 & & Simple & Buildings & 50.3 & 50.3 & 50.0 \\ 
 & & Simple & Maps & 68.4 & 68.4 & 68.4 \\ 
 & & Instr. & Landscapes & 60.9 & 60.6 & 56.0 \\ 
 & & Instr. & Animals & 70.6 & 63.6 & 62.6 \\ 
 & & Instr. & Artworks & 75.5 & 67.1 & 61.8 \\ 
 & & Instr. & Buildings & 50.5 & 50.5 & 50.5 \\ 
 & & Instr. & Maps & 52.3 & 54.7 & 52.5 \\ 
\cmidrule(lr){2-7}
 & \multirow{10}{*}{3-Shot}   & Simple & Landscapes & 66.85 & 67.03 & 66.93 \\ 
 & & Simple & Animals & 58.14 & 52.18 & 44.59 \\ 
 & & Simple & Artworks & 73.81 & 69.39 & 66.02 \\ 
 & & Simple & Buildings & 52.16 & 52.91 & 50.91 \\ 
 & & Simple & Maps & 50.79 & 53.68 & 44.77 \\ 
 & & Instr. & Landscapes & 69.00 & 70.38 & 68.47 \\ 
 & & Instr. & Animals & 78.86 & 75.78 & 76.40 \\ 
 & & Instr. & Artworks & 71.98 & 68.28 & 65.11 \\ 
 & & Instr. & Buildings & 47.61 & 46.64 & 42.70 \\ 
 & & Instr. & Maps & 52.40 & 58.95 & 50.17 \\ 
\bottomrule
\end{tabular}
    \caption{Direct VQA judging per Category: We present macro Precision (P), Recall (R), and F1 against human annotations for the combined judgments of \textbf{Q2} and \textbf{Q3} (full prompt).}
\end{table}

\begin{table}[h]
    \centering
\begin{tabular}{llllccc}
\toprule
    \textbf{Model} &  \textbf{Strategy} & \textbf{Prompt} & \textbf{Category} & \textbf{P} &  \textbf{R} & \textbf{F1} \\
\midrule
 \multirow{1}{*}{Baseline} && & Overall & 48.4 & 48.3 & 47.7\\
\midrule
\multirow{20}{*}{GPT-4o mini} & \multirow{10}{*}{0-Shot}   & Simple & Landscapes & 59.15 & 56.74 & 47.83 \\ 
 & & Simple & Animals & 76.27 & 63.75 & 61.97 \\ 
 & & Simple & Artworks & 70.73 & 70.81 & 70.76 \\ 
 & & Simple & Buildings & 57.51 & 54.15 & 53.93 \\ 
 & & Simple & Maps & 52.62 & 59.47 & 50.76 \\ 
 & & Instr. & Landscapes & 60.64 & 58.32 & 50.16 \\ 
 & & Instr. & Animals & 75.76 & 58.66 & 53.94 \\ 
 & & Instr. & Artworks & 74.27 & 73.33 & 71.90 \\ 
 & & Instr. & Buildings & 47.47 & 48.08 & 47.44 \\ 
 & & Instr. & Maps & 51.49 & 56.32 & 47.38 \\ 
 
\cmidrule(lr){2-7}
 & \multirow{10}{*}{3-Shot}   & Simple & Landscapes & 63.42 & 62.53 & 56.89 \\ 
 & & Simple & Animals & 69.66 & 56.61 & 51.31 \\ 
 & & Simple & Artworks & 76.29 & 71.41 & 68.08 \\ 
 & & Simple & Buildings & 55.68 & 52.63 & 51.64 \\ 
 & & Simple & Maps & 51.19 & 55.26 & 46.32 \\ 

 & & Instr. & Landscapes & 60.12 & 58.56 & 51.52 \\ 
 & & Instr. & Animals & 62.71 & 51.52 & 41.22 \\ 
 & & Instr. & Artworks & 73.67 & 73.13 & 71.96 \\ 
 & & Instr. & Buildings & 68.36 & 56.75 & 56.86 \\ 
 & & Instr. & Maps & 48.09 & 40.53 & 39.27 \\ 
\midrule
\multirow{20}{*}{GPT-5} & \multirow{10}{*}{0-Shot}  & Simple & Landscapes & 67.72 & 68.08 & 63.99 \\ 
 & & Simple & Animals & 54.17 & 50.66 & 40.71 \\ 
 & & Simple & Artworks & 72.11 & 72.32 & 71.96 \\ 
 & & Simple & Buildings & 89.69 & 56.52 & 55.79 \\ 
 & & Simple & Maps & 47.45 & 48.95 & 48.19 \\ 
 & & Instr. & Landscapes & 68.06 & 67.85 & 63.00 \\ 
 & & Instr. & Animals & 71.11 & 57.80 & 53.22 \\ 
 & & Instr. & Artworks & 78.79 & 78.79 & 78.00 \\ 
 & & Instr. & Buildings & 88.89 & 52.17 & 47.92 \\ 
 & & Instr. & Maps & 47.45 & 48.95 & 48.19 \\ 

\cmidrule(lr){2-7}
 & \multirow{10}{*}{3-Shot}  & Simple & Landscapes & 76.95 & 78.49 & 74.88 \\ 
 & & Simple & Animals & 53.56 & 52.42 & 50.18 \\ 
 & & Simple & Artworks & 76.40 & 76.57 & 75.99 \\ 
 & & Simple & Buildings & 90.10 & 58.70 & 59.32 \\ 
 & & Simple & Maps & 47.45 & 48.95 & 48.19 \\ 
 & & Instr. & Landscapes & 69.23 & 69.43 & 65.00 \\ 
 & & Instr. & Animals & 71.82 & 63.22 & 61.90 \\ 
 & & Instr. & Artworks & 73.94 & 73.84 & 73.00 \\ 
 & & Instr. & Buildings & 89.69 & 56.52 & 55.79 \\ 
 & & Instr. & Maps & 47.40 & 47.89 & 47.64 \\ 
\bottomrule
\end{tabular}
    \caption{Direct VQA judging per Category: We present macro Precision (P), Recall (R), and F1 against human annotations for the combined judgments of \textbf{Q2} and \textbf{Q3} (full prompt).}
\end{table}

\clearpage

\subsection{Additional Results on Model Errors}

\begin{figure}[h]
    \centering
    \includegraphics[width=0.7\linewidth]{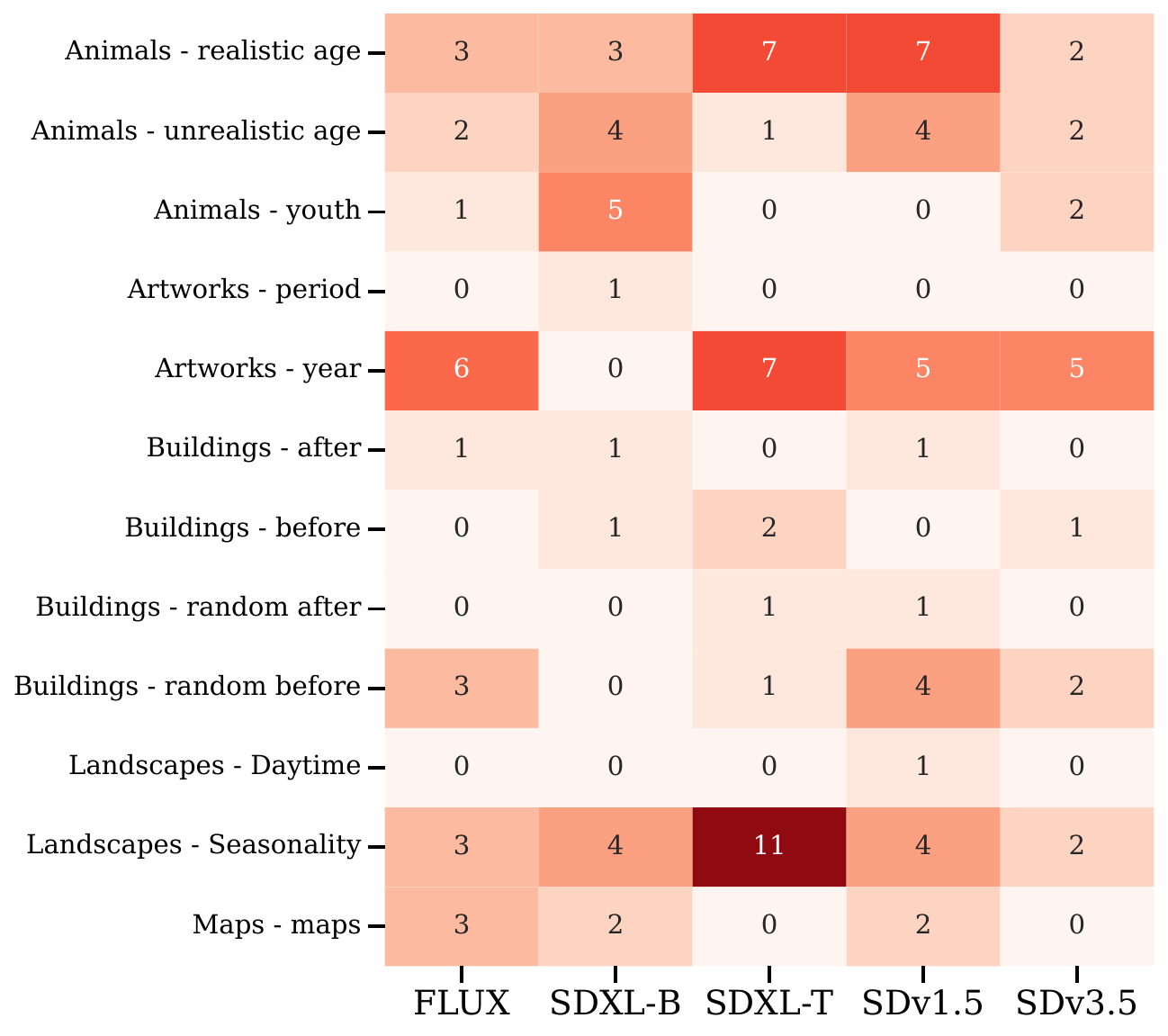}
    \caption{GPT-5 error distribution across temporal categories and T2I models for the best performing prompting strategy.}
    \vspace{-1em}
\end{figure}

\end{document}